\documentclass[runningheads]{llncs}
\usepackage{subcaption} 
\usepackage{amsmath} 

\usepackage{graphicx}
\begin{document}
\title{Impact of Face Alignment \\on Face Image Quality}
%
%
\author{Eren Onaran\inst{1}\orcidID{0009-0003-7369-4023} \and Erdi Sarıtaş\inst{1}\orcidID{0009-0001-0493-6792} \and Hazım Kemal Ekenel\inst{1,2}\orcidID{0000-0003-3697-8548}}

\institute{Department of Computer Engineering, Istanbul Technical University \and Division of Engineering, NYU Abu Dhabi \\
\email{\{onarane21,saritas21,ekenel\}@itu.edu.tr\\he2244@nyu.edu}
}
\maketitle              
\begin{abstract}

Face alignment is a crucial step in preparing face images for feature extraction in facial analysis tasks. For applications such as face recognition, facial expression recognition, and facial attribute classification, alignment is widely utilized during both training and inference to standardize the positions of key landmarks in the face. It is well known that the application and method of face alignment significantly affect the performance of facial analysis models. However, the impact of alignment on face image quality has not been thoroughly investigated. Current FIQA studies often assume alignment as a prerequisite but do not explicitly evaluate how alignment affects quality metrics, especially with the advent of modern deep learning-based detectors that integrate detection and landmark localization. To address this need, our study examines the impact of face alignment on face image quality scores. We conducted experiments on the LFW, IJB-B, and SCFace datasets, employing MTCNN and RetinaFace models for face detection and alignment. To evaluate face image quality, we utilized several assessment methods, including SER-FIQ, FaceQAN, DifFIQA, and SDD-FIQA. Our analysis included examining quality score distributions for the LFW and IJB-B datasets and analyzing average quality scores at varying distances in the SCFace dataset. Our findings reveal that face image quality assessment methods are sensitive to alignment. Moreover, this sensitivity increases under challenging real-life conditions, highlighting the importance of evaluating alignment’s role in quality assessment.

\keywords{Face alignment \and Face image quality assessment.}
\end{abstract}
\section{Introduction}

Face image analysis is one of the most actively studied fields in biometrics~\cite{FR-Survey,FR-Survey2,FIQA-Survey,fac-survey,fer-survey}. This field branches into multiple subfields, each focused on extracting meaningful features for specific applications, often under real-life conditions. Some of the most commonly studied areas are face recognitions~\cite{FR-Survey,FR-Survey2}, face image quality assessments~\cite{FIQA-Survey}, facial expression recognition~\cite{fer-survey}, and facial attribute classification~\cite{fac-survey}. 
However, the performance of these applications is heavily influenced by the conditions under which face images are acquired.

While factors like resolution, noise, and sharpness are often used to assess image quality, they can sometimes be misleading. These general metrics fail to consider face-specific challenges such as pose, occlusion, and illumination. To address these limitations, specialized approaches for assessing the quality of face images have been developed~\cite{FIQA-Survey}.

Face Image Quality Assessment (FIQA) aims to map a face image to a quality score, supporting downstream tasks—most notably face recognition. FIQA plays a critical role in determining whether a face image is suitable for processing~\cite{FIQA-Survey} and ensures that image quality is effectively accounted for in biometric systems.

Between face detection and feature extraction, an essential step conventionally used in most facial analysis tasks, including Face Image Quality Assessment (FIQA), is face alignment. Face alignment involves locating predefined landmarks on the face, such as the eyes' centers, the nose's tip, and the mouth's corners. A geometric transformation is computed using these landmarks to ensure that the detected facial features are positioned consistently across all images in the dataset. This standardization results in the uniform relative positioning of facial features~\cite{FR-Survey,FR-Survey2,hazımhocaalignment1,hazımhocaalignment2}.

Modern deep learning-based face detectors further facilitate alignment by integrating facial landmark detection into their pipelines. Models like MTCNN~\cite{MTCNN}, RetinaFace~\cite{RetinaFace}, STN~\cite{STN}, and CenterFace~\cite{CenterFace} have advanced the alignment process, improving both accuracy and robustness in real-world scenarios.

Although face alignment is a common preprocessing step in facial analysis, its broader impact on subsequent tasks, such as quality assessment, remains insufficiently explored. To address this need, our study investigates the effect of alignment on face image quality by comparing the quality scores of cropped and aligned face images. To assess face image quality, we employed several methods, including SER-FIQ~\cite{SER-FIQ}, FaceQAN~\cite{FaceQAN}, DifFIQA~\cite{DifFIQA}, and SDD-FIQA~\cite{SDD-FIQA}. For detection and alignment, we used MTCNN~\cite{MTCNN} and RetinaFace~\cite{RetinaFace} models across the LFW~\cite{LFW}, IJB-B~\cite{IJB-B}, and SCFace~\cite{SCFace} benchmark datasets.

To analyze the effect of alignment, we generated two versions of each face image for both detection models: one cropped and one aligned. The cropped images were obtained based on the bounding box coordinates provided by the detection models, while the aligned images were generated using the located facial landmarks. We then calculated the quality scores for both the ``Cropped" and ``Aligned" images to enable a comparative analysis. Our findings reveal that misalignment consistently reduces face image quality across a variety of conditions. This effect particularly stands out in challenging scenarios, such as surveillance, where images are often of lower quality.

This paper is organized as follows. Section~\ref{literatür} gives necessary background information and briefly reviews recent advancements. Section~\ref{method} explains the methodology and describes the tools used. Section~\ref{experiments} details our experimental setup. Section~\ref{results} discusses our findings. Finally, we present our conclusions in Section~\ref{conclusion}.

\section{Background} \label{literatür}

This section presents background information and reviews recent advancements in the field to provide the necessary context.

\subsection{Face Detection \& Alignment}

Face detection involves identifying faces within an image and providing their bounding box coordinates, while face alignment focuses on locating predefined landmarks on the detected faces~\cite{FR-Survey,FR-Survey2}. These landmarks typically include five key points: the corners of the eyes, the tip of the nose, and the corners of the mouth~\cite{MTCNN,RetinaFace,STN,CenterFace}.

Recent advancements in deep learning have significantly improved these processes by leveraging the inherent correlation between detection and landmark localization through joint learning methods such as MTCNN~\cite{MTCNN}, RetinaFace~\cite{RetinaFace}, STN~\cite{STN}, and CenterFace~\cite{CenterFace}. These models simultaneously detect faces and predict key facial landmarks, enabling more accurate and efficient alignment. Using these landmarks, face images can be geometrically transformed. In some cases, affine or perspective transformations are applied to correct variations in the angle or tilt of the face, ensuring consistent alignment.

Accurately locating facial landmarks is increasingly difficult under challenging conditions such as poor lighting, extreme poses, occlusions, or low-resolution images. These conditions obscure or distort key facial features, leading to
errors in the alignment process, causing misaligned face images and ultimately impacting facial analysis~\cite{align-bias,align-bias2}.

\subsection{Face Image Quality Assessment}

FIQA methods assign quality scores to evaluate how suitable a face image is for processing, leveraging statistical analysis, machine learning, and deep learning models~\cite{FIQA-Survey}. The quality scores produced by a FIQA method are influenced by factors such as lighting, blur, brightness, contrast, pose, expression, occlusion, resolution, and noise~\cite{fiqa-bias-ass}. Due to variations in training methodologies, datasets, and model architectures, different FIQA models respond uniquely to these factors. Consequently, the definition of face image quality is inherently tied to its utility for a specific task and model rather than conforming to a universal definition of image quality.

With quality scores defined based on their utility, FIQA is utilized in various applications, particularly in face recognition systems. In this context, FIQA provides value in several key areas: filtering out low-quality images at the time of capture, optimizing database maintenance by ranking and selecting high-quality images, and calibrating recognition systems to manage variations in image quality~\cite{FIQA-Survey}. FIQA helps reduce false negatives and enhances overall system performance by identifying potentially poor quality images before processing.

Recent FIQA methods are predominantly based on deep learning techniques. These methods can be categorized into two main types: model-based methods~\cite{SER-FIQ,FaceQAN,DifFIQA,SDD-FIQA,pfe,grafiqs} and regression-based methods~\cite{DifFIQA,faceqnet,pcnet}. Model-based methods employ a face recognition model to operate directly in the embedding space to calculate quality scores. On the other hand, regression-based methods generate pseudo labels for quality scores and then train neural networks using these labels.

Model-based FIQA methods leverage embedding-level properties to evaluate the quality of face images by utilizing a face recognition model. Unlike regression-based methods, these approaches assess image quality directly within the feature space of a face recognition model. PFE (Probabilistic Face Embeddings)~\cite{pfe} models face embeddings as probability distributions, producing both a mean vector (embedding) and a variance vector (uncertainty). The quality score is derived from the variance, with lower uncertainty reflecting higher image quality. SER-FIQ~\cite{SER-FIQ}, an unsupervised method, evaluates face image quality by measuring the consistency of embeddings generated with stochastic dropout. FaceQAN~\cite{FaceQAN} integrates using adversarial perturbations to calculate the pseudo quality scores. DifFIQA~\cite{DifFIQA} employs denoising diffusion probabilistic models (DDPMs)~\cite{DDPM}, perturbing facial images through the forward and backward processes of DDPMs to quantify the impact on embeddings. GraFIQs~\cite{grafiqs} also considers the use of gradients through a face recognition model. To calculate the gradients, they measure the differences in the statistics of Batch Normalization layers.

Regression-based FIQA methods focus on generating informative pseudo-labels for quality scores. FaceQNet~\cite{faceqnet} selects an anchor image for each individual using third-party ICAO compliance software. Next, pseudo quality score labels are calculated as the cosine similarity between the anchor image and the other images of the same individual. PCNet~\cite{pcnet} leverages a large number of mated image pairs (distinct images of the same individual), deriving quality labels from the embedding similarities of these pairs. Similarly, SDD-FIQA~\cite{SDD-FIQA} uses inter-class and intra-class distances to compute quality scores as the average Wasserstein distance between mated and non-mated comparison distributions. The DifFIQA~\cite{DifFIQA} can be categorized as both regression-based and model-based. Its main contribution is in the model-based scope, where it also provides a regression model due to the inference time.

\section{Methodology} \label{method}

We employed MTCNN~\cite{MTCNN} and RetinaFace~\cite{RetinaFace} for face detection and landmark localization. For assessing the quality of the faces, we used four quality assessment models: SER-FIQ~\cite{SER-FIQ} (based on the ArcFace model~\cite{ArcFace}), FaceQAN~\cite{FaceQAN}, DifFIQA~\cite{DifFIQA}, and SDD-FIQA~\cite{SDD-FIQA}. 

\subsection{Face Detectors}
\begin{itemize}
    
    \item MTCNN~\cite{MTCNN}: The Multi-task Cascaded Convolutional Network (MTCNN) is a popular face detection method with three stages that progressively improve face localization. The process starts with the Proposal Network (P-Net), which scans the image to identify candidate face windows. Due to its broad search, the network often captures many false positives at this stage. Next comes the Refine Network (R-Net), which takes these initial candidate windows and applies more precise filtering, removing most false positives and keeping only the areas most likely to contain faces. In the final stage, the Output Network (O-Net) enhances the results further, refining the remaining windows and adding precise bounding boxes with five facial landmarks (two eyes, the nose, and two mouth corners) to support alignment and boost detection accuracy. The cascaded approach helps MTCNN balance efficiency and accuracy, making it well-suited to detect faces in various poses and lighting conditions. 

    \item RetinaFace~\cite{RetinaFace}: RetinaFace is a robust face detection model built on the RetinaNet~\cite{RetinaNet} object detection framework and enhanced with advanced multi-task learning techniques for optimized performance. With a Feature Pyramid Network (FPN) backbone, RetinaFace includes context modules at each feature level, allowing it to capture rich contextual details and improve detection accuracy across diverse face sizes and poses. Unlike traditional two-stage face detection models, RetinaFace operates in a single stage, delivering both speed and precision. For each detected face, RetinaFace provides a bounding box, a confidence score, dense 3D face vertices, and five key facial landmarks (eye corners, nose, and mouth corners) to support precise alignment. This unified approach enables RetinaFace to excel at face localization and alignment, even in challenging, real-world scenarios like occlusions, varied lighting, and extreme poses.
\end{itemize}

\subsection{Face Image Quality Assessment}

\begin{itemize}
    
    \item SER-FIQ~\cite{SER-FIQ}: The Stochastic Embedding Robustness for Face Image Quality (SER-FIQ) method assesses the quality of a face image by examining how consistent the feature embeddings are across multiple forward passes with dropout applied. This approach evaluates image quality based on the stability of embeddings. An image is considered high quality if its embeddings remain consistent across multiple passes through the model. Conversely, when there is a large variance between these embeddings, it indicates higher uncertainty, pointing to a lower-quality image. SER-FIQ measures this by calculating the negative mean Euclidean distance between the embeddings generated in each pass and then applies a sigmoid function to standardize the score. 
    
    \item FaceQAN~\cite{FaceQAN}: Face Quality Assessment Network (FaceQAN) assesses the quality of an image by examining how well its feature embeddings withstand adversarial noise, giving insight into its resilience in challenging conditions. The process starts by calculating the cosine similarity between the embeddings of the original face image and a noise-injected version. Using the Fast Gradient Sign Method (FGSM)~\cite{FSGM}, the model then applies adversarial noise to the original image over multiple iterations, observing how this affects the embeddings. A quality score is calculated by combining the mean and variance of these cosine similarity scores and the similarity score between the original image and its horizontally flipped version. By focusing on consistency and robustness in this way, FaceQAN provides a quality score where higher values indicate superior stability under adversarial conditions.
    
    \item DifFIQA~\cite{DifFIQA}: Denoising Diffusion Face Image Quality Assessment (DifFIQA) measures face image quality by utilizing both the forward and backward steps of diffusion processes. The method starts by gradually degrading the image through several forward diffusion steps, adding noise in each step. Then, it reconstructs the image using a Denoising Diffusion Probabilistic Model (DDPM)~\cite{DDPM} during the backward steps. This process is also applied to a horizontally flipped version of the image, resulting in six variations of the original image. These variations are then passed through a face recognition model to extract feature embeddings. Then, cosine similarities are calculated between the embeddings of the original image and each modified version. This process is repeated over multiple iterations, with the average cosine similarity across all iterations as the final quality score. A regression model is trained using these quality scores as pseudo labels to speed up the assessment process, allowing quick quality evaluation without repeated diffusion steps. In the experiments, a regression model pre-trained with pseudo labels is used.
    
    \item SDD-FIQA~\cite{SDD-FIQA}: Similarity Distribution Distance Face Image Quality Assessment (SDD-FIQA) method evaluates face image quality by analyzing how well an image’s embedding can differentiate between samples of the same identity (intra-class) and those of different identities (inter-class). Ideally, high-quality images are close to others of the same identity while clearly separated from different identities. The process begins by creating a positive set (same identity) and a negative set (different identities) for each query image, selecting challenging examples through a hard sampling strategy. Cosine similarity scores between the query image and images in each set are then computed, forming positive and negative distributions. The Wasserstein distance is calculated to measure the separation between these distributions, serving as a pseudo-label that represents image quality. A regression model is then trained on these pseudo-labels, allowing quick quality assessments based on distribution distance—where a greater separation typically indicates higher image quality.

\end{itemize}

\subsection{Face Alignment}

Initially, the detection models estimated bounding boxes and five facial landmarks: the left eye, right eye, nose tip, left mouth corner, and right mouth corner. No scaling is applied for the bounding boxes provided by RetinaFace, and the coordinates are used directly. In contrast, for faces detected using MTCNN, the bounding boxes are scaled by a factor of 1.2 to address their tightness, and the faces are cropped accordingly. The datasets generated from these cropped face images, based on MTCNN and RetinaFace, are referred to as ``Cropped".

\begin{figure}[!t]
    \centering
    \includegraphics[width=0.75\linewidth]{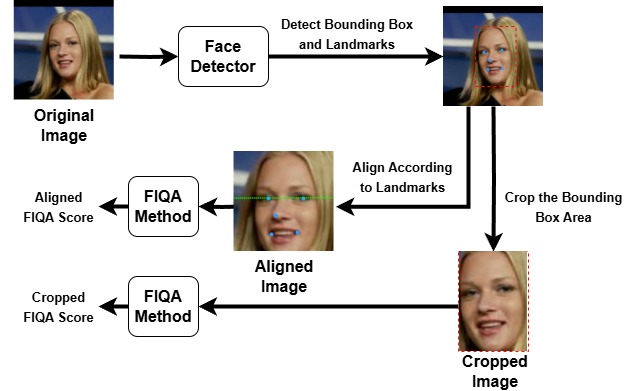}
    \caption{The general overview of our pipeline is as follows. A face detector first processes an original image containing a face. The detector outputs a bounding box (red dotted frame) and facial landmarks (blue dots). The image is then cropped to extract the face region, resulting in what we refer to as the ``Cropped Image." In parallel, the input image is aligned based on the detected landmarks, producing the ``Aligned Image." This alignment standardizes the face image, such as ensuring the line connecting the eyes (green dotted line) is parallel to the horizontal axis. Finally, both the cropped and aligned images are evaluated using a FIQA method to estimate their FIQA scores.}
    \label{fig:pipeline}
\end{figure}

The face images are also processed for alignment using the detected facial landmarks. Each of the five landmarks is mapped to a pre-determined position in the output image based on the template used in ArcFace~\cite{ArcFace}. A similarity transformation matrix is computed to minimize the sum of squared distances between the transformed landmark positions and their corresponding target locations. Notably, this alignment process is applied directly to the original images and operates independently of the cropping operation.

\begin{equation}
    T = \underset{T}{\arg\min} \sum_{i \in \text{landmarks}} \| T(p_i) - q_i \|^2
\end{equation}
where $|| . ||$ represents the $L_2$ distance, $p_i$ denotes the detected landmark, and $q_i$ is its desired position. The transformation matrix $T$ performs three operations: scaling, rotation, and translation. Mathematically, $T$ is defined as:
\begin{equation}
    T(\begin{bmatrix} x \\ y \end{bmatrix}) = 
    \alpha \begin{bmatrix} \cos\theta & -\sin\theta \\ \sin\theta & \cos\theta \end{bmatrix}\begin{bmatrix} x \\ y \end{bmatrix}
    + \begin{bmatrix} t_x \\ t_y \end{bmatrix}
\end{equation}
which can also be expressed in homogeneous coordinates as:
\begin{equation}
T(\begin{bmatrix} x \\ y \\ 1 \end{bmatrix}) = 
\begin{bmatrix} \alpha\cos\theta & -\alpha\sin\theta & t_x\\ \alpha\sin\theta & \alpha\cos\theta & t_y \\ 0 & 0 & 1\end{bmatrix}\begin{bmatrix} x \\ y \\ 1 \end{bmatrix}
\end{equation}
Here, $\alpha$ is the scaling factor that ensures the distance between the eyes remains fixed, $\theta$ is the rotation angle that aligns the line connecting the eyes to the horizontal axis, and $t_x$, $t_y$ are the translation parameters used to standardize the face location. The transformation $T$ is applied to the input image to warp it for alignment. Datasets created from these aligned face images are referred to as ``Aligned". The overall pipeline is illustrated in Fig.~\ref{fig:pipeline}.

\section{Experimental Setup} \label{experiments}

In this study, we utilized three face image datasets: LFW~\cite{LFW}, IJB-B~\cite{IJB-B}, and SCFace~\cite{SCFace}.

\subsection{Datasets}

\begin{itemize}
    
    \item Labeled Faces in the Wild (LFW)~\cite{LFW}: LFW is a widely recognized benchmark dataset for evaluating face recognition and verification models in unconstrained settings. It includes 13,233 images of 5,749 individuals collected from the internet, with variations in pose, expression, lighting, background, and occlusions. Its primary evaluation procedure consists of 6,000 face image pairs divided into ten folds, with each fold having 300 positive and 300 negative pairs. Then, 10-fold cross-validation is applied; for each fold, the nine other folds are used to find an optimal threshold, and the current fold is used to calculate the test performance.
          
    \item The IARPA Janus Benchmark-B (IJB-B)~\cite{IJB-B}:  IJB-B is a comprehensive benchmark dataset designed for unconstrained face recognition. It consists of 11,754 images and 7,011 videos featuring 1,845 subjects. The dataset covers a range of variations, including pose and quality, with detailed annotations on occlusions, skin tone, and eye/nose location in the bounding boxes. In the evaluation procedure of IJB-B, the 1-1 verification protocol is used generally, where similarity scores are calculated between pre-determined pairs. Then, these similarity scores are used to calculate TAR@FAR (True Accept Rate at Fixed False Accept Rates) or ROC (Receiver Operating Characteristic) Curves.

    \item Surveillance Cameras Face Database (SCFace)~\cite{SCFace}: SCFace is a specialized benchmark dataset created for surveillance applications. It features images of 130 subjects captured at three distances (4.2m, 2.6m, and 1.0m) using five different video surveillance cameras of varying quality, resulting in 15 images per subject. Each subject also has a standard mugshot image. For each combination of distance and camera, captured face images are compared with mugshot images, and the identity of the closest match is assigned. For each distance, the accuracy across all camera settings is averaged to evaluate performance.
    
\end{itemize}

\begin{figure}[!b]
    \centering
    \includegraphics[width=0.4\linewidth]{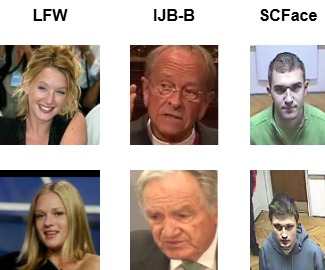}
    \caption{Sample original images from LFW~\cite{LFW}, IJB-B~\cite{IJB-B}, and SCFace~\cite{SCFace}.}
    \label{fig:samples}
\end{figure}

\begin{figure}[!t]
    \centering
    \includegraphics[width=0.85\linewidth]{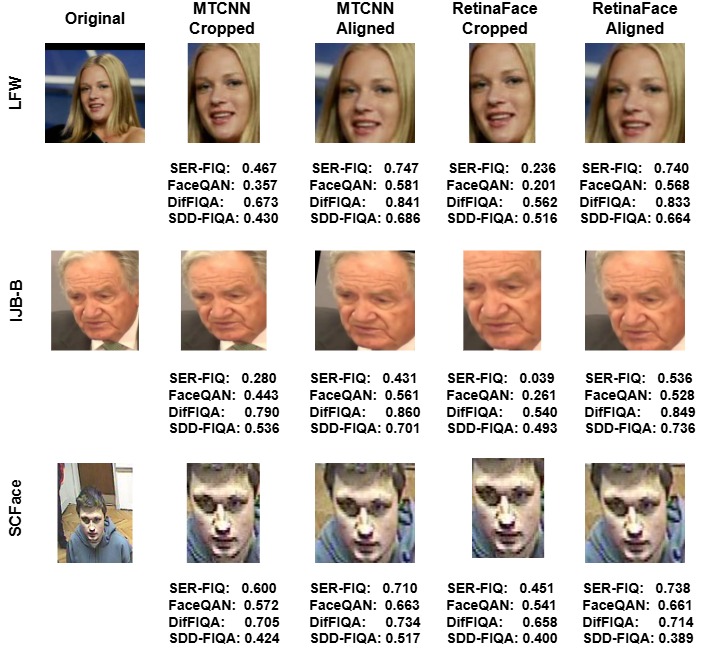}
    \caption{Sample original images from LFW~\cite{LFW}, IJB-B~\cite{IJB-B}, and SCFace~\cite{SCFace}, their cropped and aligned versions. The results of each FIQA method are written below in each cropped and aligned image.}
    \label{fig:samplesWscores}
\end{figure}

Sample original images from each dataset are shown in Fig.~\ref{fig:samples}. For each original image, the bounding box and landmark coordinates are determined using MTCNN and RetinaFace. Then, ``Cropped" and ``Aligned" versions of these images are generated. Quality scores for each version are then computed using the specified FIQA methods. Sample outputs are displayed in Fig.~\ref{fig:samplesWscores}, where the original images are presented in the first column. The corresponding detector and whether the image is cropped or aligned are indicated at the top of each column, while the FIQA scores for each version are provided below.

\section{Results} \label{results}

We began our analysis by first investigating the input image variations and individual FIQA results shown in Fig.~\ref{fig:samplesWscores}. Examining the quality scores, we find that the cropped versions consistently score lower across all examples, reflecting the negative impact of misalignment. Moreover, this effect is more apparent when the original images are more misaligned.

\begin{figure}[!t]
\begin{subfigure}{.5\textwidth}
  \centering
  \includegraphics[width=1.05\linewidth]{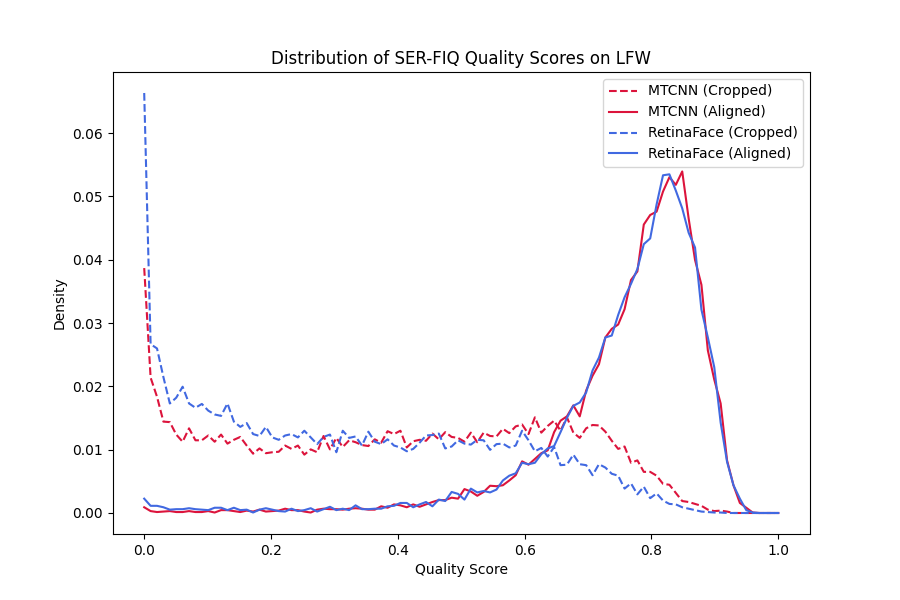}
  \caption{SER-FIQ}
  \label{fig:lfw:serfiq}
\end{subfigure}
\hspace{-0.025\linewidth}
 \begin{subfigure}{.5\textwidth}
  \centering
  \includegraphics[width=1.05\linewidth]{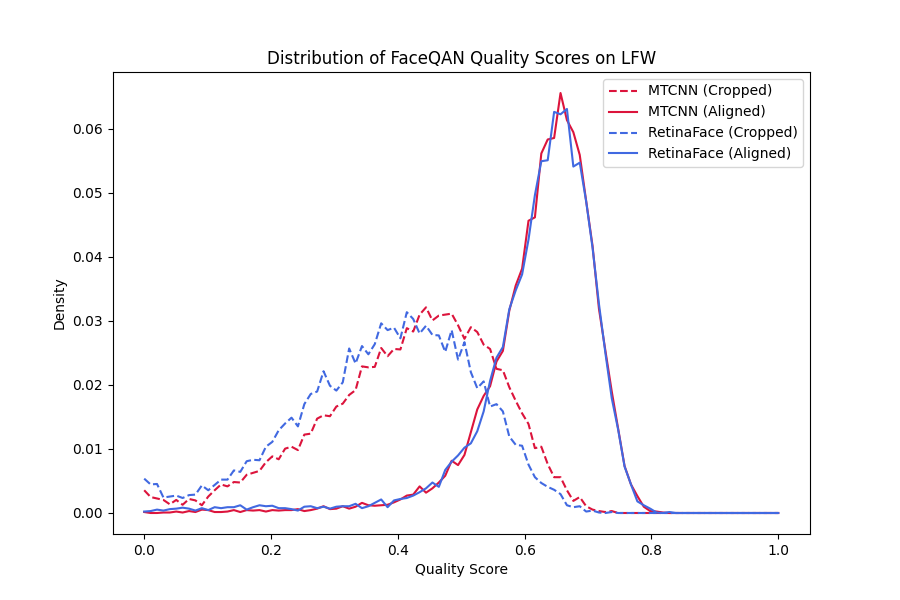}
  \caption{FaceQAN}
  \label{fig:lfw:faceqan}
\end{subfigure}

 \begin{subfigure}{.5\textwidth}
  \centering
  \includegraphics[width=1.05\linewidth]{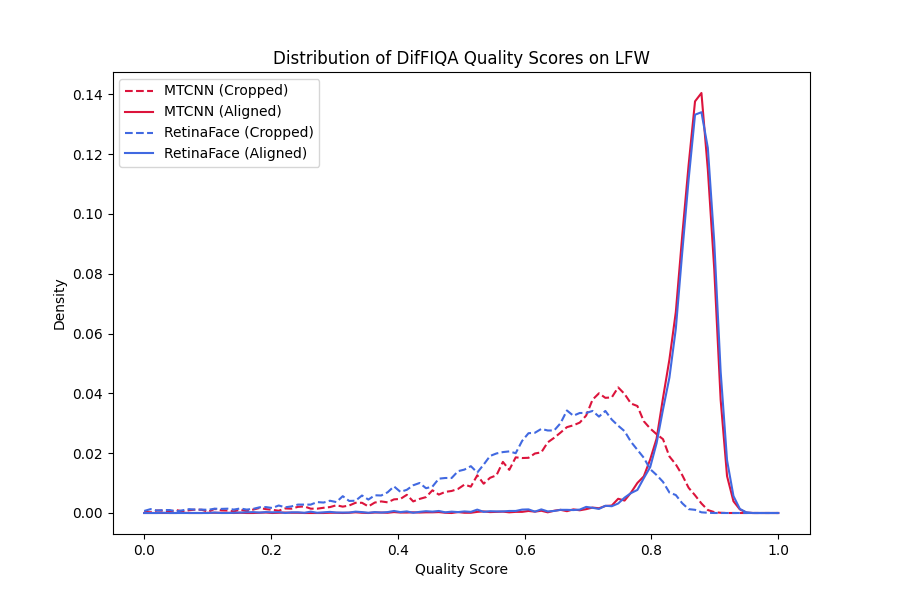}
  \caption{DifFIQA}
  \label{fig:lfw:diffiqa}
\end{subfigure}
\hspace{-0.025\linewidth}
 \begin{subfigure}{.5\textwidth}
  \centering
  \includegraphics[width=1.05\linewidth]{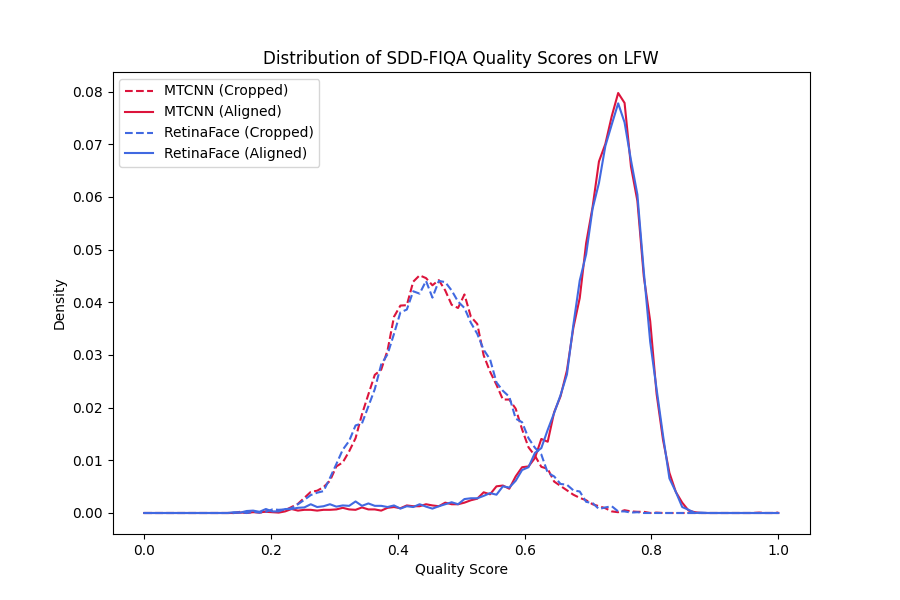}
  \caption{SDD-FIQA}
  \label{fig:lfw:sddfiqa}
\end{subfigure}
 
  \caption{Quality score distributions for LFW~\cite{LFW} dataset. }  
  \label{fig:lfw}
\end{figure}

\begin{figure}[!t]
\begin{subfigure}{.5\textwidth}
  \centering
  \includegraphics[width=1.05\linewidth]{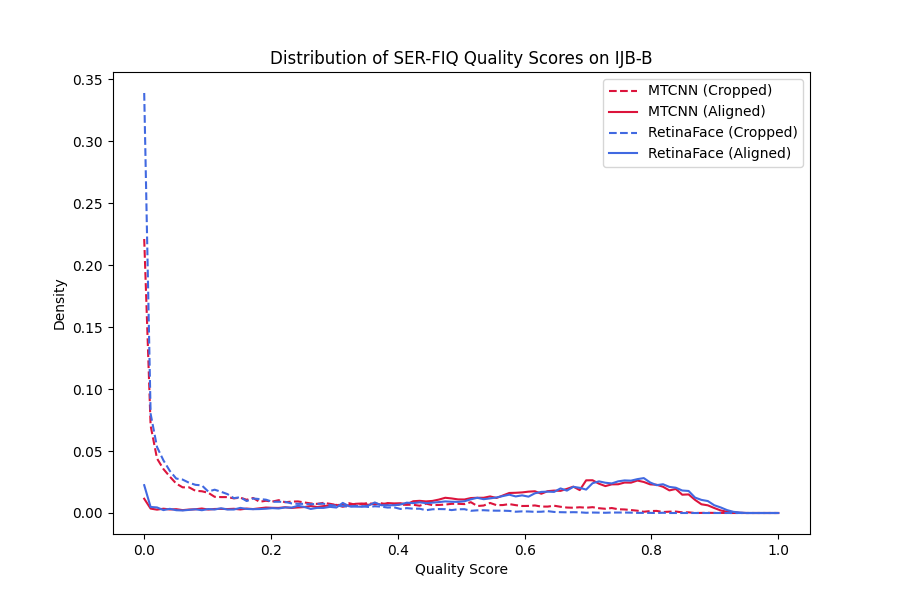}
  \caption{SER-FIQ}
  \label{fig:ijbb:serfiq}
\end{subfigure}
\hspace{-0.025\linewidth}
 \begin{subfigure}{.5\textwidth}
  \centering
  \includegraphics[width=1.05\linewidth]{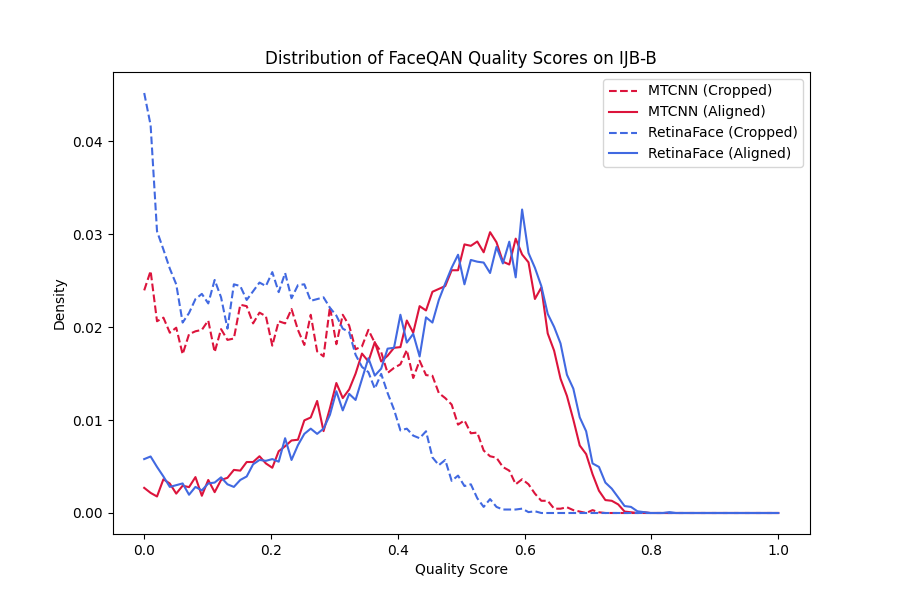}
  \caption{FaceQAN}
  \label{fig:ijbb:faceqan}
\end{subfigure}

 \begin{subfigure}{.5\textwidth}
  \centering
  \includegraphics[width=1.05\linewidth]{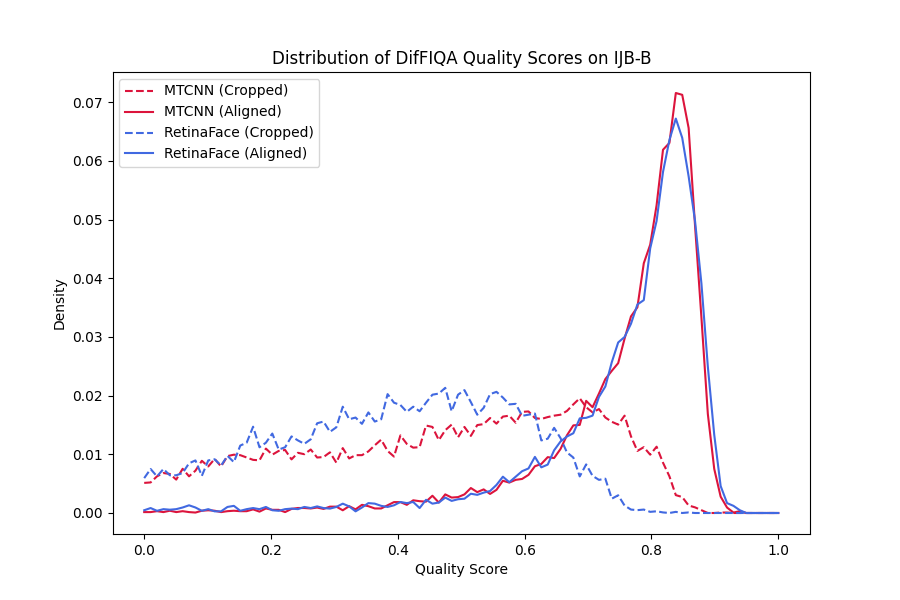}
  \caption{DifFIQA}
  \label{fig:ijbb:diffiqa}
\end{subfigure}
\hspace{-0.025\linewidth}
 \begin{subfigure}{.5\textwidth}
  \centering
  \includegraphics[width=1.05\linewidth]{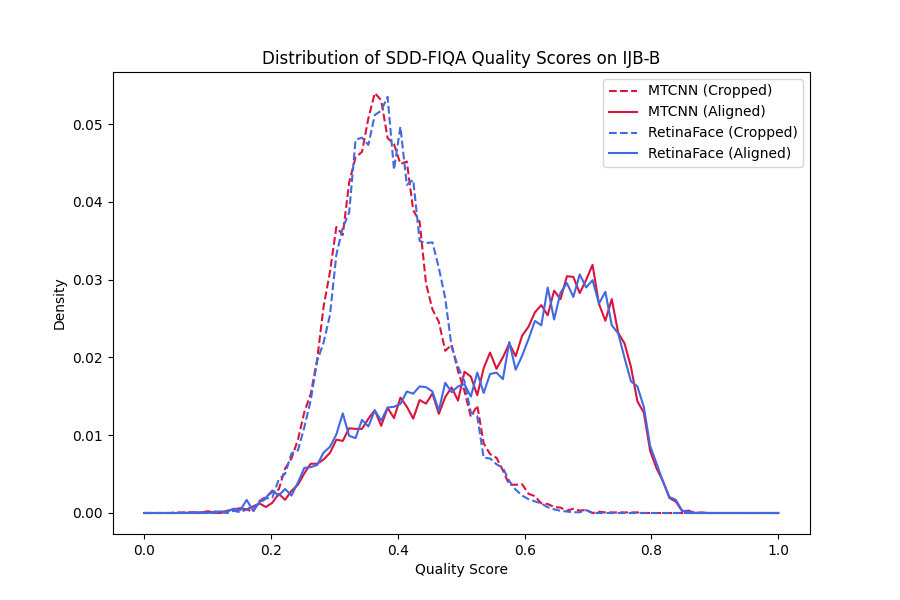}
  \caption{SDD-FIQA}
  \label{fig:ijbb:sddfiqa}
\end{subfigure}
   \caption{Quality score distributions for IJB-B~\cite{IJB-B} dataset. }

  \label{fig:ijbb}
\end{figure}

Quality score distributions for the LFW and IJB-B datasets are presented in Fig.~\ref{fig:lfw} and Fig.~\ref{fig:ijbb}, respectively. Each figure includes four subplots, corresponding to the outputs of different FIQA methods: SER-FIQ~\cite{SER-FIQ} (top-left), FaceQAN~\cite{FaceQAN} (top-right), DifFIQA~\cite{DifFIQA} (bottom-left), and SDD-FIQA~\cite{SDD-FIQA} (bottom-right). In each subplot, the y-axis represents density, while the x-axis shows the quality scores. The plots feature four results: MTCNN~\cite{MTCNN} cropped (red dotted line), MTCNN aligned (solid red line), RetinaFace~\cite{RetinaFace} cropped (blue dotted line), and RetinaFace aligned (solid blue line).

For the LFW results in Fig.~\ref{fig:lfw}, the quality score distributions of aligned images have higher averages and lower variances than the ones of cropped images. The quality score distributions of aligned images are very similar, whether MTCNN or RetinaFace is used for landmark localization, while cropped images show slight variations in their distributions. In the SER-FIQ results (Fig.~\ref{fig:lfw:serfiq}), the quality scores have a high density near zero, followed by a nearly uniform distribution up to 0.6. In contrast, the quality scores for aligned images follow a Gaussian distribution with averages close to 0.8. For the remaining plots—Fig.~\ref{fig:lfw:faceqan}, Fig.~\ref{fig:lfw:diffiqa}, and Fig.~\ref{fig:lfw:sddfiqa}—the quality score distributions for both cropped and aligned images exhibit approximate Gaussian shapes. The results can be summarized based on the means of the quality score distributions. FaceQAN's mean quality score is approximately 0.45 for cropped images and 0.7 for aligned images. For DifFIQA, the mean quality score is around 0.7 for cropped images and 0.85 for aligned images. For SDD-FIQA, the mean score is approximately 0.45 for cropped images and 0.75 for aligned images. 

In the IJB-B results shown in Fig.~\ref{fig:ijbb}, the quality score distributions for aligned images are nearly identical, while the distributions for cropped images show more notable distinctions. For the SER-FIQ results in Fig.~\ref{fig:ijbb:serfiq}, the density of quality scores near zero is much higher compared to the one in Fig.~\ref{fig:lfw:serfiq}, and the scores beyond this range exhibit a more flat, low-density distribution. Moreover, for the aligned images, we can see Gaussian distributions with the means close to 0.75. For FaceQAN (Fig.~\ref{fig:ijbb:faceqan}), the quality scores for cropped images are notably lower and look like a monotonically decreasing function. In contrast, the quality score distributions of aligned images form a curved shape with means close to 0.55. For the DifFIQA results (Fig.~\ref{fig:ijbb:diffiqa}), the quality score distribution for cropped images has a mean near 0.5 and high variance, while the quality score distribution for aligned images has a mean near 0.8 and low variance. For SDD-FIQA (Fig.~\ref{fig:ijbb:sddfiqa}), the distributions for cropped images show lower variances than those for aligned images. The average quality scores are approximately 0.4 for cropped images and around 0.7 for aligned images. These results from two datasets -- Fig.~\ref{fig:lfw} and Fig.~\ref{fig:ijbb} -- clearly show that cropped images consistently achieve lower mean quality scores compared to aligned images.

\begin{table}[!b]
    \caption{SCFace~\cite{SCFace} dataset results.}
    \label{tab:scface}
    \centering
\resizebox{0.80\columnwidth}{!}{%
    \begin{tabular}{|c|c|c|c|c|c|}
    \hline
        \textbf{FIQA}  & ~ & \multicolumn{2}{c|}{\textbf{MTCNN}} & \multicolumn{2}{c|}{\textbf{RetinaFace}}  \\ \cline{3-6}
        \textbf{Method} & \textbf{Distance} & \textbf{Cropped} & \textbf{Aligned} & \textbf{Cropped} & \textbf{Aligned} \\ \hline
        ~                   & 1.0m      & 0.600& \textbf{0.789}& 0.448& \textbf{0.792}\\ \cline{2-6}
        SER-FIQ             & 2.6m      & 0.483& \textbf{0.705}& 0.285& \textbf{0.711}\\ \cline{2-6}
        ~                   & 4.2m      & 0.187& \textbf{0.474}& 0.070& \textbf{0.426}\\ \hline
        ~                   & 1.0m      & 0.548& \textbf{0.665}& 0.475& \textbf{0.666}\\ \cline{2-6}
        FaceQAN             & 2.6m      & 0.497& \textbf{0.586}& 0.396& \textbf{0.588}\\ \cline{2-6}
        ~                   & 4.2m      & 0.331& \textbf{0.451}& 0.198& \textbf{0.407}\\ \hline
        ~                   & 1.0m      & 0.712& \textbf{0.800}& 0.633& \textbf{0.802}\\ \cline{2-6}
        DifFIQA             & 2.6m      & 0.630& \textbf{0.715}& 0.531& \textbf{0.710}\\ \cline{2-6}
        ~                   & 4.2m      & 0.461& \textbf{0.578}& 0.316& \textbf{0.543}\\ \hline
        ~                   & 1.0m      & 0.419& \textbf{0.584}& 0.413& \textbf{0.579}\\ \cline{2-6}
        SDD-FIQA            & 2.6m      & 0.371& \textbf{0.435}& 0.360& \textbf{0.422}\\ \cline{2-6}
        ~                   & 4.2m      & 0.331& \textbf{0.342}& 0.325& \textbf{0.331}\\ \hline
    \end{tabular}
    }%
\end{table}

The SCFace results are shown in Table~\ref{tab:scface}. The table presents the average quality scores for each FIQA method at specific distances across different scenarios, including the detection method and whether the images are aligned or cropped. Consistent with previous findings, aligned face images exhibit higher mean quality scores than cropped face images. Additionally, the quality scores decrease as the distance at which the image is captured increases. The contribution to face image quality is most pronounced in the SER-FIQ results, followed by FaceQAN and DifFIQA. Among the FIQA methods, SDD-FIQA shows the least sensitivity to alignment by producing close quality scores between cropped and aligned images. The results indicate that alignment is a crucial step for FIQA in surveillance.

\subsection{Discussion}

Faces are often captured upright, making them almost naturally aligned after cropping with a well-predicted bounding box. However, our results clearly show that using only cropped images produces inferior quality scores. Similar to face recognition~\cite{align-bias,align-bias2}, this behaviour may result from the FIQA methods being trained with aligned images. However, this much quality drop when using the cropped images still remains unexpected. The results demonstrate that the FIQA methods are sensitive to alignment.

The images in the selected datasets were collected under real-world conditions. Therefore, they contain various factors that degrade image quality, such as distance, lighting, or occlusions. Moreover, when these factors are combined, their impact amplifies each other~\cite{combineddeg}, resulting in the loss of vital facial details. These degradation factors can lead to faulty localization of landmarks, which can result in misalignment. We can draw an indirect relation between alignment and face image quality; as the quality of images becomes lower, it is harder to estimate the correct positions of the landmarks, resulting in misalignment and, in turn, lower quality scores. 

\section{Conclusion} \label{conclusion}

Face alignment is crucial for preparing face images for feature extraction and significantly influences face recognition performance~\cite{hazımhocaalignment1,hazımhocaalignment2}. However, how the alignment affects the face image quality has not been explored in detail. In this study, we assessed the impact of face alignment by utilizing quality scores from SER-FIQ, FaceQAN, DifFIQA, and SDD-FIQA. We conducted our analysis on the benchmark datasets, including LFW, IJB-B, and SCFace.  For bounding box and facial landmark localization, we used the MTCNN and RetinaFace models. To analyze the impact of the alignment on image quality, we compared the quality scores of cropped images with the aligned ones.

We analyzed quality score distributions on the LFW and IJB-B datasets and reviewed average quality scores at different distances in the SCFace dataset. Misalignment had a major negative effect on quality scores in all the datasets. Even though this quality score reduction is significant for relatively normal real-world condition datasets like LFW and IJB-B, the effect is more apparent in challenging scenarios like surveillance, as evidenced by the results of the SCFace dataset, especially at greater capturing distances. Thus, the alignment process could be more crucial for surveillance scenarios where face images are likely low quality. Our findings indicate that misaligned face images -- only cropped versions -- produce lower quality scores than properly aligned ones.

It is essential to remember that the FIQA methods were trained using aligned face images. Even though image quality may not directly relate to misalignment, the results show that FIQA methods are sensitive to misaligned face images. 
For some facial analysis applications, it might be desirable to minimize the quality score variations of these models when dealing with faces that are not ideally aligned.

\section*{ACKNOWLEDGMENTS}
\small This research is supported by the TÜBİTAK 2247-C Intern Researcher Scholarship Programme (STAR) and the Meetween Project, funded by the European Union’s Horizon Europe Research and Innovation Programme under Grant Agreement No. 101135798.

\bibliographystyle{splncs04}
\bibliography{bibliography}

\end{document}